\documentclass[mlabstract]{jmlr}

\newcommand{\R}{\mathbb{R}}

\newcommand{\E}{\mathbb{E}}

\newcommand{\I}{\mathbf{I}}
\newcommand{\1}{\mathbf{1}}

\newcommand{\M}{\mathcal{M}}

\newcommand{\br}[1]{\left(#1\right)}

\newcommand{\cat}{\text{Cat}}

\usepackage{longtable}% for long tables

\usepackage{booktabs}
\usepackage[load-configurations=version-1]{siunitx} % newer version
\theorembodyfont{\upshape}
\theoremheaderfont{\scshape}
\theorempostheader{:}
\theoremsep{\newline}

\jmlrvolume{}
\firstpageno{1}
\jmlryear{2025}
\title{The Cosine Schedule is Fisher-Rao-Optimal for Masked Discrete Diffusion Models}

\author{
    \Name{Leo Zhang} \Email{leo.zhang@stx.ox.ac.uk} \\ \addr University of Oxford 
    \AND
    \Name{Saifuddin Syed} \Email{saif.syed@stat.ubc.ca} \\ \addr University of British Columbia
}

\begin{document}

\maketitle

\begin{abstract}
In this work, we study the problem of choosing the discretisation schedule for sampling from masked discrete diffusion models in terms of the information geometry of the induced probability path.
Specifically, we show that the optimal schedule under the Fisher-Rao geometry recovers the popularly-used cosine schedule.
\end{abstract}
\begin{keywords}
Discrete Diffusion Models, Information Geometry.
\end{keywords}

\section{Introduction}
\label{sec:intro}

% Restore CHANGED for submission
Modern generative models, such as diffusion and flow-based models \citep{song2020score, lipman2022flow, albergo2023stochastic}, generate samples via the numerical simulation of dynamical processes.
For instance, sampling from a continuous diffusion model requires simulating the time-reversal SDE parametrised by the learnt score function. 
Therefore, the choice of discretisation schedule is crucial for generating high quality samples.

While the problem of choosing an optimal discretisation schedule has been well-studied in the context of continuous generative models \citep{watson2021learning, santos2023using, sabour2024align, williams2024score}, there has been growing interest \citep{park2024jump} in this question for discrete diffusion \citep{austin2021structured, campbell2022continuous, shi2024simplified, sahoo2024simple}.
Here, the dynamical process underlying this class of models is a continuous-time Markov chain (CTMC) \citep{del2017stochastic} constructed from some forward corruption process of (discrete) data; samples are then generated by simulating a learnt approximation to the time-reversal CTMC.
Common choices of the forward corruption process include masked discrete diffusion \citep{shi2024simplified, sahoo2024simple} and uniform discrete diffusion \citep{sahoo2025diffusion}.

\looseness=-1
In this work, we study the choice of discretisation schedule for masked discrete diffusion models from an information-geometric perspective \citep{amari2016information}.
In particular, we show that the probabilistic structure of masked discrete diffusion allows for the closed-form computation of the Fisher-Rao metric of the induced probability path.
In turn, this allows us to derive the optimal schedule---in terms of the Fisher-Rao geometry of the path---as the geodesic under this geometry, following the approach of \cite{syed2022non, williams2024score, syed2024optimised}.
Interestingly, we find that the optimal schedule recovers to the popularly-used cosine schedule \citep{nichol2021improved}.

\section{Background}

\paragraph{Masked discrete diffusion} 
Using the notation of \cite{shi2024simplified}, we consider the masked stochastic process $x_t = (x_t^{(1)}, \ldots, x_t^{(N)})$ with $t\in[0, 1]$ and $N\in\mathbb{N}$, where each $x_t^{(n)}$ represents a discrete token in $[m]=\{0, \ldots, m\}$; in particular, we take $m$ to denote the special ``masked'' state.
The forward noising process is then given by:
\[
q(x_t|x_0) = \prod_{n=1}^N q(x_t^{(n)}|x_0^{(n)}), \quad q(x_t^{(n)}|x_0^{(n)}) = \cat\br{x_t^{(n)}; \bar{Q}(t)^\top x_0^{(n)}},
\]
where
\[
\bar{Q}(t) = \alpha_t\I + (1-\alpha_t)\1e_m^\top, \quad \alpha_t = \exp\br{-\int_0^t\beta(s)\, ds},
\]
for some continuous function $\beta:[0, 1]\to\R^+$ which ensures that $\alpha_0=1$ and $\alpha_1\approx0$.
We denote the probability path induced by this forward process as $q_t(x_t) = \E_{q_{x_0}(x_0)}[q(x_t|x_0)]$, where $q_0$ is our initial data distribution.

\paragraph{Information geometry}
A parametrised collection of probability densities $\M=\{p_\theta: \theta\in\Theta\subset\R^d\}$ can be geometrically studied as a Riemannian manifold \citep{amari2016information} by endowing the parameter space $\Theta$ with a Riemannian metric $\delta(\theta)\in\R^{d\times d}$ \citep{lee2018introduction}.
One natural choice is the Fisher-Rao metric $I(\theta)\in\R^{d\times d}$ given by:
\[
    I(\theta)_{ij} = \E_{X\sim p_\theta}\left[ \frac{\partial}{\partial\theta_i}\log p_\theta(X) \frac{\partial}{\partial\theta_j}\log p_\theta(X) \right].
\]
We can view the Fisher-Rao metric as characterising the geometry induced by the Kullback–Leibler (KL) divergence from the Taylor expansion \citep{polyanskiy2025information}:
\[
    \text{KL}(p_{\theta_0}||p_{\theta_0+\Delta\theta}) = \frac{1}{2}\Delta\theta^\top I(\theta)\Delta\theta + O(\Delta\theta^3).
\]
This framework has been used to study generative models in \cite{das2023image, santos2023using, Perone, williams2024score, karczewski2025spacetime}.

\paragraph{Optimal schedules}
We note that choosing an discretisation schedule $\mathcal{T}=\{t_i\}_{i=0}^T$ where  $0=t_0<t_1<\ldots<t_T=1$ can be viewed as a reparametrisation of time.
Formally, let $\varphi:[0, 1]\to[0, 1]$ be a strictly increasing, differentiable function such that $\varphi(0)=0$ and $\varphi(1)=1$. 
We say $\mathcal{T}$ is generated by $\varphi$ if $t_i=\varphi(i/T)$.
From an information-geometric perspective, we can view $\varphi$ as a regular path in the manifold $\M = \{q_t: t\in[0, 1]\}$.

In order to provide a notion of optimality, \cite{syed2022non, williams2024score, syed2024optimised} considers placing a metric $\delta(t)\in\R$ on $\M$ so that the geometric length of the path between neighbouring discretisation points $q_{t_i}, q_{t_{i+1}}$ captures a suitable notion of the ``cost'' from transitioning between $q_{t_i}$ and $q_{t_{i+1}}$\footnote{These works respectively choose metrics in the context of parallel tempering (a MCMC algorithm) and continuous diffusion models.}.
In particular, the length $\Lambda$ of the entire path is defined in the standard way as
\[
    \Lambda = \int_0^1 \sqrt{\delta(\varphi(t))}\Dot{\varphi}(t) dt.
\]

Hence, given some choice of metric on $\M$, the generator $\varphi^*$ for the optimal schedule can then be naturally defined as the geodesic between $q_0$ and $q_1$.
By a standard calculation, the form of $\varphi^*$ is given by Theorem \ref{prop:geodesics} below. See Appendix \ref{proof:geodesics} for the proof.
\begin{theorem}\label{prop:geodesics}
    The optimal schedule under the metric $\delta(t)\in\R$ is generated by $\varphi^*$ of the form:
    \[
        \varphi^*(t) = \Lambda^{-1}(\Lambda t) \ \text{ where } \ \Lambda(s) = \int_0^s \sqrt{\delta(r)} dr.
    \]
\end{theorem}
This formula is due to the fact that geodesics not only minimise the length of the path\footnote{We note that since the manifold is 1D and we assume $\varphi$ is a regular path, the length is actually invariant to different choices of $\varphi$.} between $q_0$ and $q_1$ but traverse the path at a constant rate. 
For example, we see that the length between $q_{t_i^*}$ and $q_{t_{i+1}^*}$ equals $\Lambda/T$ for $t_i^* = \varphi^*(i/T)$.
Therefore, the optimal schedule ensures that the cost between discretisation points is equally spaced apart.

\paragraph{Intuition behind the optimal schedule}
In the context of sampling from a diffusion model, the optimal schedule corresponds to discretisation steps that transition between marginals $q_t$ that have uniform discrepancy between them (as measured by the chosen metric).
This avoids taking large steps in the regions where $q_t$ is rapidly changing and small steps where $q_t$ is changing slowly, optimising our computational budget.
Further, we note that the Fisher-Rao metric is a sensible, generic choice in this context due to the aforementioned intimate connection with the KL divergence.

\section{Main Result}

\looseness=-1
Below we provide a statement of our main result. See Appendix \ref{proof:optimal-schedule} for the proof.
\begin{theorem}\label{thm:optimal-schedule}
    The optimal schedule $\mathcal{T}^*=\{t_i^*\}_{i=0}^T$ where $0=t_0^*<t_1^*<\ldots<t_N^*=1$ for a masked discrete diffusion model under the Fisher-Rao metric satisfies:
    \[
        \alpha_{t_i^*} = \cos^2\br{ \frac{i}{N}\cdot\br{\frac{\pi}{2}-\arcsin\sqrt{\alpha_1}}}.
    \]
    The values of $t_i^*$ are given by the formula: $t_i^* = F^{-1}(-\log\alpha_{t_i^*})$ where $F(t)=\int_0^t\beta(s) ds$.
\end{theorem}
As a trivial consequence of this, we have the following corollary.
\begin{corollary}
    For a choice of $\beta$ such that $\alpha_1=0$ (e.g. $\beta(s)=\frac{1}{1-s}$ and $\alpha_t=1-t$ using the convention that $\exp(-\infty)=0$), the optimal schedule under the Fisher-Rao metric satisfies:
    \[
        \alpha_{t_i^*} = \cos^2\br{\frac{i}{T}\cdot\frac{\pi}{2}}.
    \]
\end{corollary}
We see that this schedule coincides with the cosine schedule heuristically introduced in \cite{nichol2021improved} (with $s=0$) for continuous diffusion models.
A similar result was proved by \cite{santos2023using} which showed for continuous diffusion model, when the data distribution $q_0(x_0)$ is a Dirac measure---i.e. $q_0(x_0) = \delta_{x}(x_0)$, the cosine schedule is also optimal under the Fisher-Rao geometry.
Another related result can be found in Theorem 6 from \cite{chehab2023provable} which shows that the Fisher-Rao geodesic between some distributions $p_0$ and $p_1$ converges to the arithmetic annealing path with the cosine schedule, as the support of $p_0, p_1$ become more disjoint.
These results provide an interesting theoretical justification for the use of the cosine schedule.

\section{Conclusion}
We have shown that the optimal schedule under the Fisher-Rao geometry for masked discrete diffusion models recovers the popularly-used cosine schedule; we leave to future work the empirical validation of this schedule choice.
Moreover, we hope this spurs further investigation into the information-geometric properties of generative models, for instance, considering optimal schedules under different metric choices.
Finally, one limitation of our work is that we only consider the properties of the true probability path $q_t$. 
This does not take into account either the errors in the approximation of $q_t$ or the sampling scheme \citep{kim2025train} used during generation.
It is an interesting future direction to study the information geometry of these settings.

\newpage

% CHANGED
\acks{
 LZ and SS thank Peter Potaptchik for helpful conversations. 
 LZ is supported by the EPSRC CDT in Modern Statistics and Statistical Machine Learning (EP/S023151/1).
 SS acknowledges support from the NSERC Postdoctoral Fellowship.
 }

\bibliography{pmlr-sample}

\newpage

\appendix

\section{Proofs}

\subsection{Proof of Theorem \ref{prop:geodesics}}\label{proof:geodesics}

\begin{proof}
    We recall from standard Riemannian geometry the definition of the energy $E(\varphi)$ of the path $\varphi$:
    \[
        E(\varphi) = \int_0^1 \delta(\varphi(s)) \Dot{\varphi}(s)^2 ds.
    \]
    We note that while the length in our context is invariant for all suitable choices of $\varphi$, the energy is not invariant - i.e. the energy depends on the rate at which we traverse the path.
    It is a standard result that the geodesic will minimise the energy functional; we will reproduce this proof for completeness.

    By the Cauchy–Schwarz inequality applied to functions in the space $L^2([0, 1])$, we have:
    \begin{align*}
        \Lambda^2 = \left(\int_0^1 \sqrt{\delta(\varphi(s))}\Dot{\varphi}(s) ds\right)^2 &= \left\langle 1, \sqrt{\delta(\varphi(s))}\Dot{\varphi}(s) \right\rangle_{L^2([0, 1])}^2 \\
        &\le ||1||^2_{L^2([0, 1])} \cdot ||\sqrt{\delta(\varphi(s))}\Dot{\varphi}(s)||_{L^2([0, 1])}^2 \\
        &= \int_0^1 \delta(\varphi(t))\Dot{\varphi}(s)^2 ds \\
        &= E(\varphi).
    \end{align*}
    It is a standard result that this inequality becomes an equality if and only if $1$ and $\sqrt{\delta(\varphi(s))}\Dot{\varphi}(s)$ are linearly dependent.
    Therefore, taking $\sqrt{\delta(\varphi^*(s))}\Dot{\varphi}^*(s) = C$
    for some constant $C>0$, we have
    \[
        Ct = \int_0^t \sqrt{\delta(\varphi^*(s))}\Dot{\varphi}^*(s) ds = \int_0^{\varphi^*(t)} \sqrt{\delta(r)} dr = \Lambda(\varphi^*(t)),
    \]
    where $C=\Lambda$ from taking $t=1$ and the second equality follows from the substitution $r=\varphi^*(t)$. Moreover, this implies that
    \[
        \varphi^*(t) = \Lambda^{-1}(\Lambda t).
    \]
    We see by construction that $\varphi^*$ is both a length minimiser and travels at a constant rate, hence, we can conclude that $\varphi^*$ is a geodesic.
\end{proof}

\subsection{Proof of Theorem \ref{thm:optimal-schedule}}\label{proof:optimal-schedule}

\begin{proof}
\looseness=-1
Due to the structure of the probability path induced by masked discrete diffusion models, we can derive the following analytic form for $q_t(x_t)$:
\[
q_t(x_t) = (1-\alpha_t)^{n_m(x_t)}\alpha_t^{N-n_m(x_t)}\E_{q(x_0)}\left[\1(x_t^{(i)} = x_0^{(i)}: x_t^{(i)}\ne m ) \right],
\]
and by a similar argument, $\partial_t q_t(x_t)$ has the form:
\begin{align*}
\partial_t q_t(x_t) 
&= \partial_t \E_{q(x_0)}\left[ \prod_{n=1}^N q(x_t^{(i)}|x_0^{(i)}) \right] \\
&= \partial_t (1-\alpha_t)^{n_m(x_t)}\alpha_t^{N-n_m(x_t)}\E_{q(x_0)}\left[\1(x_t^{(i)} = x_0^{(i)}: x_t^{(i)}\ne m ) \right],
\end{align*}
where we define $n_m(x_t)$ as the number of masked tokens $m$ present in $x_t$.

Hence, the Fisher score has the form:
\begin{align*}
    \partial_t\log q_t(x_t) 
    &= \partial_t \log \br{ (1-\alpha_t)^{n_m(x_t)}\alpha_t^{N-n_m(x_t)} } \\
    &= - \frac{n_m(x_t)\Dot{\alpha}_t}{1-\alpha_t} + (N - n_m(x_t))\frac{\Dot{\alpha}_t}{\alpha_t} .
\end{align*}
To compute the Fisher-Rao metric $I(t)\in\R$, we can use the standard result: 
\[
    I(t) = \text{Var}_{x_t\sim q_t}\left(\partial_t\log q_t(x_t)\right),
\]
with the fact that $n_m(x_t)$ follows a $\text{Bin}(N, 1-\alpha_t)$ distribution when $x_t\sim q_t$ (since the probability of $x_t^{(i)}$ being a masked token $m$ is Bernoulli and we assume a factorised forward process). 
This allows us to conclude
\begin{align*}
    I(t) 
    = \text{Var}_{x_t\sim q_t}(\partial_t\log q_t(x_t))
    &= \frac{N\alpha_t(1-\alpha_t)\Dot{\alpha}_t^2}{\alpha_t^2(1-\alpha_t)^2} \\
    &= \frac{N\Dot{\alpha}_t^2}{\alpha_t(1-\alpha_t)}.
\end{align*}
In order to apply Theorem \ref{prop:geodesics}, we compute:
\begin{align*}
    \Lambda(t) = \int_0^t \sqrt{\frac{N\Dot{\alpha}_t^2}{\alpha_t(1-\alpha_t)}} dt 
    = -\int_{1}^{\alpha_t} \sqrt{\frac{N}{y(1-y)}} dy 
    &= 2\sqrt{N}[\arcsin\sqrt{y}]_{\alpha_t}^1  \\
    &= 2\sqrt{N}\left(\frac{\pi}{2} - \arcsin\sqrt{\alpha_t}\right),
\end{align*}
where for the first equality, we use the substitution $y=\alpha_t$ and $\frac{|\Dot{\alpha}_t|}{\Dot{\alpha}_t}=-1$ since $\alpha_t$ is constructed to be strictly decreasing, and for the second equality, we use a standard integral identity.

Furthermore, to find the generator $\varphi^*$ for the optimal schedule, we use the identity $\Lambda(\varphi^*(t)) = \Lambda t$ from Theorem \ref{prop:geodesics} to show
\begin{align*}
    2\sqrt{N}\left( \frac{\pi}{2} - \arcsin\sqrt{\alpha_{\varphi^*(t)}} \right) &= 2\sqrt{N}t\left( \frac{\pi}{2} - \arcsin\sqrt{\alpha_1} \right) \\
    \implies \alpha_{\varphi^*(t)} &= \sin^2\left( \frac{\pi}{2} - t\left( \frac{\pi}{2} - \arcsin\sqrt{\alpha_1} \right) \right) \\
    &= \cos^2\left( t\cdot\br{ \frac{\pi}{2} - \arcsin\sqrt{\alpha_1} } \right),
\end{align*}
where the last equality uses the trigonometric identity $\sin(a+b)=\sin a\cos b + \sin b\cos a$.
We conclude the main result from substituting the values $t=i/T$ into $\varphi^*$.
Finally, the formula for $t_i^*$ is a simple consequence of the fact that $t\mapsto\alpha_t$ is an injective map.

\end{proof}

\end{document}